\documentclass[runningheads]{llncs}
\usepackage[T1]{fontenc}

\usepackage{makeidx}  
\usepackage{mathtools}
\usepackage{enumitem}
\usepackage{hyperref}
\usepackage[capitalise]{cleveref}
\usepackage{amsfonts}
\usepackage{algorithm}
\usepackage[noend]{algpseudocode}
\usepackage{epsfig}
\usepackage{svg}
\usepackage{amsmath}
\usepackage{hyperref}
\usepackage{tabularx}
\usepackage{verbatim}
\usepackage{booktabs}
\usepackage[scaled]{helvet}
\usepackage{tikz}
\usetikzlibrary{arrows, shapes, automata, positioning}
\usepackage{caption}
\usepackage{subcaption}
\usepackage{adjustbox}
\usepackage{comment}

\newcommand{\RomanNumeralCaps}[1]
    {\MakeUppercase{\romannumeral #1}}

\begin{document}
%
%
\title{A Universal Approach to Feature Representation in Dynamic Task Assignment Problems}
\titlerunning{Feature Representation in Assignment Problems}  
%

\author{Riccardo Lo Bianco\inst{1, 2} \and
Remco Dijkman\inst{1, 2} \and
Wim Nuijten\inst{1, 2} \and
Willem van Jaarsveld\inst{1,2}\
}

\authorrunning{Riccardo Lo Bianco et al.}   
%
\tocauthor{Riccardo Lo Bianco}
\institute{$^1$ Eindhoven University of Technology, Netherlands\newline
$^2$ Eindhoven Artificial Intelligence Systems Institute, Netherlands\newline
\email{\{r.lo.bianco|r.m.dijkman|w.p.m.nuijten|w.l.v.jaarsveld\}@tue.nl}
}
\maketitle

\begin{abstract}  Dynamic task assignment concerns the optimal assignment of resources to tasks in a business process. Recently, Deep Reinforcement Learning (DRL) has been proposed as the state of the art for solving assignment problems. DRL methods usually employ a neural network (NN) as an approximator for the policy function, which ingests the state of the process and outputs a valuation of the possible assignments. However, representing the state and the possible assignments so that they can serve as inputs and outputs for a policy NN remains an open challenge, especially when tasks or resources have features with an infinite number of possible values. To solve this problem, this paper proposes a method for representing and solving assignment problems with infinite state and action spaces. In doing so, it provides three contributions:
(\RomanNumeralCaps{1}) A graph-based feature representation of assignment problems, which we call assignment graph;
(\RomanNumeralCaps{2}) A mapping from marked Colored Petri Nets to assignment graphs;
(\RomanNumeralCaps{3}) An adaptation of the Proximal Policy Optimization algorithm that can learn to solve assignment problems represented through assignment graphs. To evaluate the proposed representation method, we model three archetypal assignment problems ranging from finite to infinite state and action space dimensionalities. The experiments show that the method is suitable for representing and learning close-to-optimal task assignment policies regardless of the state and action space dimensionalities.

\keywords {Business Process Optimization, Assignment Problem, Markov Decision Process, Deep Reinforcement Learning, Graph Neural Networks} \end{abstract}

\section{Introduction}
Dynamic task assignment problems, referred to in this work as assignment problems, involve optimizing the allocation of resources to tasks in a process, where new tasks (and possibly new resources) become available at runtime. In their broad context, assignment problems encompass diverse resource and task types, compatibility constraints on the possible assignments, time-related variables such as resource schedules, time-related variables, and stochastic variables. In recent years, Deep Reinforcement Learning (DRL) was proposed as the state-of-the-art solution approach for assignment problems~\cite{10.1007/978-3-031-41620-0_13,zbikowski_deep_2023,middelhuis2024learning}. Most DRL methods rely on a neural network (NN) to encode a policy function. This function takes a representation of the state of a process, called an observation, as input and produces a valuation over the possible actions, the assignments, as output.

The ideal way to tackle an assignment problem would be to model it using widely recognized notations, like BPMN or Petri Nets (PNs), and let an algorithm learn a suitable policy using observations expressed through the same notation without additional feature engineering. This is the direction taken in~\cite{10.1007/978-3-031-41620-0_13}, where a PN variant suitable to encode assignment problems is proposed, together with an algorithmic approach to train DRL models on the problems expressed through in PN form. However, in current DRL approaches, multilayer perceptrons (MLPs) remain the most commonly used neural networks for policy approximation, and this is the case for~\cite{10.1007/978-3-031-41620-0_13} as well. MLPs require fixed-size observation vectors, which makes modeling the features of observation and action vectors challenging. In particular, the feature representation problem is challenging when observation and action vectors have unknown sizes during the design phase and could potentially grow to infinity.

\begin{figure}[h!]
    \centering
    \centerline{\includegraphics[width=\textwidth]{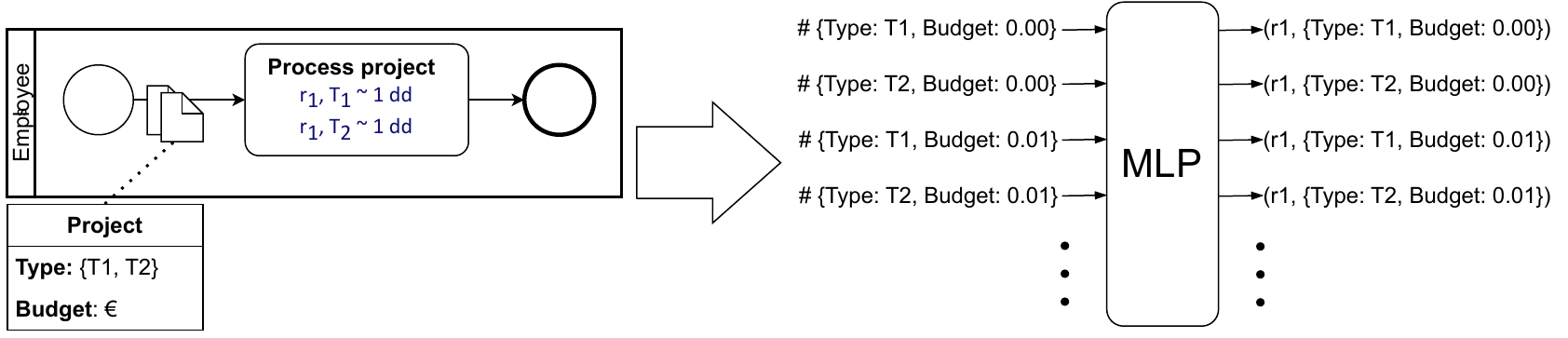}}
    \caption{Feature representation problem}
    \label{fig:representation_problem}
\end{figure}

\cref{fig:representation_problem} exemplifies this problem. On the left, it depicts a process model representing a consulting firm with a single employee, $r_1$, who can perform projects. Projects are characterized by a type, a categorical value that identifies the project type, and a budget, a value in euros representing the project's profitability. The optimization objective is to maximize profits, given by the sum of projects' budgets. On the right, the figure presents the direct translation of the process state into the observation vectors that existing DRL solution approaches use~\cite{10.1007/978-3-031-41620-0_13,zbikowski_deep_2023,middelhuis2024learning}. The elements of the input vectors are the number of projects of each possible type, while those of the output vector are the possible assignments of resource to project type (if multiple projects of the same type are queued, a FIFO strategy is applied). Since the projects are characterized by an unbounded value (the budget), the input and output sizes grow to infinity. Consequently, feature representation in vectorial form is not suitable for this problem, nor for any problem having tasks or resources characterized by continuous or unbounded variables, which encompasses many real-world problems. A prominent example in this sense is given by problems where the features depend on time, which is naturally an unbounded metric. It is possible that suitable observation vectors can be devised through feature engineering for single problem instances, but this is a tedious and error-prone task. Moreover, it may lead to observations that do not encode all information in the state space, which results in suboptimal policies being learned. For example, in the problem presented in \cref{fig:representation_problem}, the budget could be limited to a maximum value, but this choice would make the algorithm incapable of handling cases with a budget higher than the maximum. Finally, imposing a maximum budget would make the state and action spaces finite, but their dimensionalities would still grow exponentially with the number of available resources and task types in the system, which poses a significant limitation to the applicability of DRL as a solving method~\cite{tavares2018algorithms}. 

In the general case, the representation problem is twofold: on the one hand, the observation vector grows to infinity whenever continuous or unbounded variables are to be taken into account, and on the other, the action vector also grows to infinity as the result of the need to enumerate all possible assignments.

In this work, we refine the methods presented in~\cite{10.1007/978-3-031-41620-0_13} by providing three contributions:
\begin{enumerate}
    \item A graph-based feature representation approach for assignment problems, which we call assignment graph.
    \item A mapping from marked Colored Petri Nets (CPN) to assignment graphs.
    \item An adaptation of the Proximal Policy Optimization (PPO) algorithm that can learn to solve assignment problems expressed as assignment graphs.
\end{enumerate}

To evaluate the proposed method, we provide three instances of the assignment problem with increasing state and action space dimensionality. We show empirically that the novel feature representation method allows us to seamlessly learn a close-to-optimal policy on assignment problems with finite and infinite state and action spaces.

Against this background, the remainder of this paper is structured as follows. ~\cref {related_work} outlines the relevant literature. ~\cref {background} introduces the necessary background knowledge and notation. ~\cref {ae_pn} outlines the notion of assignment graph and the Petri Net to assignment graph mapping algorithm. ~\cref {ppo_graph} provides an overview of a DRL algorithm's design capable of handling dynamic assignment graphs. In~\cref{use_cases}, three assignment problem instances with increasing state and action space dimensionalities are presented, and the new feature representation method is applied to learn a close-to-optimal policy in all three cases. ~\cref{conclusion} discusses the proposed method's merits and limitations and reflects on possible directions of future research.
\section{Related work}\label{related_work}
The problem of encoding large or infinite state spaces for task assignment problems is well established in the operation research literature. Early works in this field highlighted limitations in solving variants of the fleet dispatching problem with infinite state spaces~\cite{routing_1980}.
The most widely adopted technique used to handle such problems has been aggregation. This procedure involves projecting the original problem to a state space with low dimensionality, solving the aggregated problem, and disaggregating to find the solution to the original problem~\cite{aggregation_survey}. Multiple aggregation techniques have been proposed~\cite{shape,dynamic_assignment}, but all entail arbitrary choices from the modeler or the costly tuning of ad-hoc hyperparameters.
In the field of business process optimization, assignment problems were studied through predictive process monitoring~\cite{8786063,10103533} and prescriptive process monitoring~\cite{doi:10.1080/07421222.2002.11045693,DASHTBOZORGI2023102198}. However, these methods assume predefined task and resource types, which limits their applicability to finite state spaces. In contrast, this work proposes a seamless feature representation method suitable for infinite state and action spaces.

We will frame assignment problems as a Markov Decision Process (MDP) to solve them through DRL. In this work, we will rely on the Action-Evolution Petri Net (A-E PN) framework~\cite{10.1007/978-3-031-41620-0_13}, an extension of CPN suitable to model and solve MDPs. In their original definition, A-E PNs were limited to representing finite state spaces in vectorial form as shown in~\cref{fig:example_problem}. In this work, we lift this limitation by defining a translation from a marked A-E PN to a graph, called assignment graph, that can be used as observation for a Graph Neural Network (GNN) to learn effective assignment policies.

In recent years, the use of graph-based observations has been explored as a feature representation tool for DRL applied to specific dynamic assignment problem classes, in particular in the field of transportation ~\cite{9683135,YAN2023103260,1542cc6b364e4cf8b7c2adb51da444ce}. However, these works focus on a single problem instance and do not provide a reusable approach to feature representation for assignment problems.

The use of GNNs is increasingly being explored in predictive process monitoring~\cite{10.1007/978-3-030-98581-3_9,10.1007/978-3-030-94343-1_3}. Still, the predictive setting differs significantly from decision-making, and no direct translation is possible.

Some works employed GNNs to analyze Petri Nets~\cite{sommers2021process,HU20201}. In~\cite{sommers2021process}, the close relationship between graphs and PNs is exploited to address the problem of process discovery as a supervised learning problem. In the paper, the authors encode a set of execution traces as a graph, then extract an unmarked Petri Net from the graph and train a GNN to operate deletions on the Petri Net to improve the fitness of the process model against the traces in the dataset. The paper is one of the first to employ GNNs on Petri Nets, but the algorithm is only suitable for unmarked PNs. By contrast, the technique proposed in this work enables the representation of marked A-E PNs.
In~\cite{HU20201}, the authors propose a novel GNN layer that handles a specific Petri Net variant combining timed-place Petri nets and a system of simple sequential processes with resources to optimize the dynamic scheduling problem of flexible manufacturing systems. In their work, the authors focus on defining the novel GNN layer considering a single flexible manufacturing system scheduling problem, thus tackling a smaller class of problems than ours. Moreover, the system considered does not model resources with different properties (e.g., colors) and does not deal with dynamic graphs. In the current work, we employ a general-purpose GNN layer to solve assignment problems, focusing on defining an easily adaptable translation technique from marked A-E PN to assignment graph.
\section{Background}\label{background}
Since we aim to provide a universal approach to feature representation for task assignment problems, we need to introduce a suitable modeling method and a suitable solution method. This section provides the background regarding these two components: we present a CPN variant from the literature that can be used to model assignment problems with finite state spaces. We also delineate the main steps of the PPO algorithm and the DRL algorithm we employ as a solution method.

This work expresses assignment problems using Petri Nets as a modeling language. Basic PNs are suitable to model and simulate processes of arbitrary complexity, but in their basic form, they lack the expressive power to model decision processes. In particular, PNs are not equipped with constructs to define which actions can be taken, nor the goodness of the actions taken. To bridge this gap, Action-Evolution Petri Nets were proposed~\cite{10.1007/978-3-031-41620-0_13}. A-E PNs are an extension of Timed-Arc Colored Petri Nets (T-A CPN) that serve as a modeling and solving framework for assignment problems. This definition means any assignment problem modeled through A-E PN can be directly fed to a RL algorithm to learn a close-to-optimal policy without performing feature engineering. A-E PNs distinguish two types of transitions: action transitions represent events that require the agent to make an assignment by choosing the tokens used to fire them; evolution transitions represent events that happen in the environment independently of the agent's decisions and they are fired non-deterministically. Only one type of transition can fire at a given time, based on a network tag that is possibly updated every time a transition is fired. Transitions are associated with a reward signal to account for the actions' goodness.

A reduced version of A-E PN's formal definition is reported in~\cref{def:ae_pn}. For the complete definition and the firing rules, we refer the reader to~\cite{10.1007/978-3-031-41620-0_13}.

\begin{sloppypar}
\begin{definition}[Action-Evolution Petri Net] An Action-Evolution Petri Net (A-E PN) is a tuple \(AEPN =( \mathcal{E}, P, T, F, C, G, E, I, L, l_0, \mathcal{R}, \rho_0)\), where \(\mathcal{E}\) is a finite set of types called \textit{color sets}, \(P\) is a finite set of \textit{places}, \(T\) is a finite set of \textit{transitions}, \(F\) is a finite set of \textit{arcs}, \(C : P \to \mathcal{E}\) is a \textit{color function} that maps each place into a set of possible token colors, \(G\) is a \textit{guard function}, \(E\) is an \textit{arc expression function}, \(L\) is a \textit{transition tag function}, \(I\) is an \textit{initialization function}, \(l_0\) is the \textit{network's initial tag}, which can be one of `a-transition' representing that the next step is a decision (or action) step or `e-transition', representing that the next step is a regular firing of transitions, \(\mathcal{R}\) is the \textit{transition reward function}, \(\rho_0\) is the initial \textit{network reward}. \label{def:ae_pn}
\end{definition}
\end{sloppypar}

The two elements that limit applicability of AE P-N to finite state space assignment problems are \(\mathcal{E}\) and \(C\). Consequently, we will adapt those in the next section.

\begin{figure}[h]
    \centering    \centerline{\includegraphics[width=0.7\textwidth]{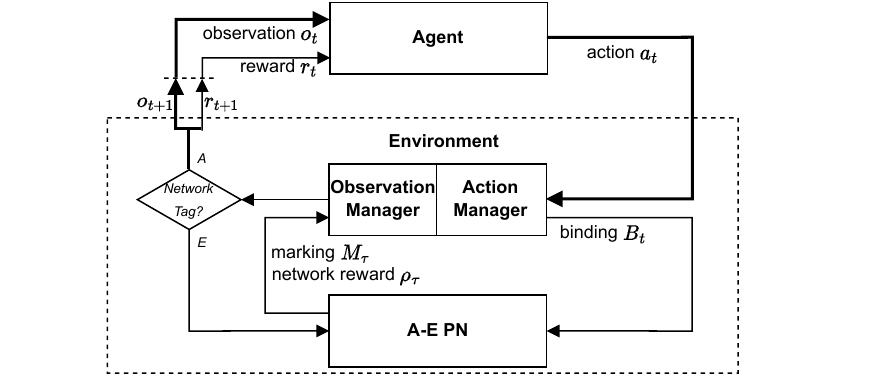}}
    \caption{The reinforcement learning cycle with A-E PN}
    \label{fig:rl_cycle_extended}
\end{figure}

~\cref{fig:rl_cycle_extended}, taken from~\cite{10.1007/978-3-031-41620-0_13}, shows how A-E PN is embedded in the RL cycle.

In the basic A-E PN framework, the observation manager is responsible for presenting a vectorized representation of the A-E PN marking to the agent, a vector containing, for each place, the number of tokens of each color in the place's color set. Similarly, the action manager is responsible for ingesting a vectorized representation of the action chosen by the agent, which is a vector having one element for each possible assignment. Both observation and action vectors are treated statically, and it is assumed that the tokens in the network fall into a limited set of types whose attributes' possible values are pre-defined by the PN color sets.
The problem of state representation in assignment problems through A-E PN arises in the same modalities presented in \cref{fig:representation_problem}.

In this work, we use a modified version of the observation manager and the action manager to handle assignment graphs of varying sizes instead of static vectors. Assignment Graphs are then fed to a specialized implementation of the PPO~\cite{schulman2017proximal} algorithm to learn close-to-optimal assignment policies.

PPO is one of the most widely adopted DRL algorithms. It is an on-policy method, which means it learns the policy used to make environmental decisions using the policy itself. We can summarize PPO's functioning mechanism as the repetition of three steps:

\begin{enumerate}
\item \textbf{Samples collection}: the agent interacts with the environment following the current policy and collects a set of trajectories. Each trajectory contains a sequence of states, actions, rewards, and new states.

\item \textbf{Advantage computation}: the agent uses the generalized advantage estimation technique~\cite{schulman2018highdimensional} to compute an estimate of how much better or worse each action was compared to the average action the current policy would have taken in that state.

\item \textbf{Policy and value function optimization}: the agent optimizes the policy to make the actions with higher advantages more likely and those with lower advantages less likely. This is done using a surrogate objective function that forces the new policy to stay close to the old policy. Parallelly, the agent updates the weights of a value network, an estimator of the total value of each encountered state based on the observed total discounted rewards at the end of trajectories to refine the computation of the advantage terms in the next steps of the optimization process.

\end{enumerate}

\section{Feature Representation through Assignment Graphs}\label{ae_pn}
This section defines a feature representation approach, namely assignment graphs, suitable for representing assignment problems with both finite and infinite state spaces, such that they can serve as input to a Neural Network. To be able to model and simulate the desired assignment problem, we first introduce a suitable PN variant. We then provide the formal definition of assignment graph and its properties. Lastly, we present a two-step mapping from a marked PN to the corresponding assignment graph.

To provide a universal approach to feature representation for assignment problems, we first need to adopt a modeling technique that is suitable to represent assignment problems with finite or infinite state space.
To this end, we provide a modified version of~\cref{def:ae_pn}, which we call attributed A-E PN.

\begin{sloppypar}
\begin{definition}[Attributed Action-Evolution Petri Net] An Attributed Action-Evolution Petri Net (A-E PN) is a tuple $AEPN =(A, P, T, F, \mathcal{A}, G, E, I, L, l_0, \mathcal{R}, \rho_0)$, where $(P, T, F, G, E, I, L, l_0, \mathcal{R}, \rho_0)$ follow~\cref{def:ae_pn}, and:
\begin{itemize}
\item \(A\) is a finite set of types called \textit{attributes}.
\item \(\mathcal{A}\) is an \textit{attribute function} that maps each place \(p\) into a set of attributes. Each token on \(p\) must have a color that is composed of the token's time and a value for each of the attributes in $\mathcal{A}(p)$.
\end{itemize}
\label{def:attributed_ae_pn}
\end{definition}
\end{sloppypar}

The main difference between A-E PN (and accordingly CPN) and attributed A-E PN is that the former expects all the places' color sets to be finite and defined at the initialization. In contrast, the latter only expects the attributes of tokens that flow in each place to be given at the initialization, while the values associated with each attribute in a given marking do not need to be pre-defined.
This work will consider cases where attributes represent values in \(\mathbb{R}\). From this point onward, we will use the term A-E PN to refer to attributed A-E PNs.

In~\cref{def:assignment_graph}, we introduce assignment graphs as a representation of assignment problems that can serve as input to a Neural Network to learn effective assignment policies. 

 \begin{definition}{\textbf{(Assignment Graph)}}
An assignment graph $G$ is a tuple $G = (V, D, Y, A, \phi, \theta)$ where: 
\begin{itemize}
 \item $V$ is a finite set of nodes.
 \item $D \subseteq V \times V$ is a set of ordered pairs of vertices, known as (directed) edges.
 \item $Y = \{ \text{A\_Transition}, \text{E\_Transition}, \dots \}$ is a finite set of node types, where $\text{A\_Transition}$ represents the action type and $\text{E\_Transition}$ represents the evolution type. Other node types represent the different places in the PN, discriminated on the base of their attributes. 
 \item $A$ is a finite set of attributes.
 \item $\phi: V \rightarrow Y$ is a function assigning a type to each node.
 \item $\theta: Y \rightarrow 2^A$ is a function assigning a set of attributes to each type.
\end{itemize} 
\label{def:assignment_graph} 
\end{definition}

An assignment graph respects the following two properties:

\begin{itemize}

\item For any two nodes $v_1, v_2 \in V$ where either $\phi(v_1) = \phi(v_2) = \text{A\_Transition}$, $\phi(v_1) = \phi(v_2) = \text{E\_Transition}$, or $\phi(v_1) = \text{A\_Transition}$ and $\phi(v_2) = \text{E\_Transition}$, there does not exist a directed edge $d \in D$ such that $d = (v_1, v_2)$ or $d = (v_2, v_1)$.

\item For any node $v \in V$ with $\phi(v) = \text{A\_Transition} \lor \phi(v) = \text{E\_Transition}$, the set of attributes assigned to it by $\theta$ is empty, i.e., $\theta(v) = \emptyset$.

\end{itemize}

\begin{figure}[h!]
    \centering
    \centerline{\includegraphics[width=\textwidth]{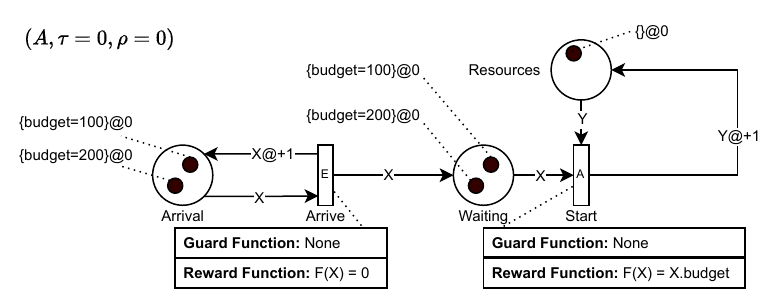}}
    \caption{Marked A-E PN for a simple consulting firm.}
    \label{fig:example_problem}
\end{figure}

Having provided the necessary definitions, we can describe the two-step algorithm to translate a marked A-E PN to the corresponding assignment graph. To clarify the algorithm's steps, we will refer to the problem of a consulting firm with a single employee in~\cref{fig:example_problem}. Tasks arrive at a constant frequency of two per time unit, characterized by a type and a budget. The resource must be allocated to tasks to maximize the total budget, considering that the completion time of an assignment is always equal to one time unit. The problem is modeled through one evolution transition (\textit{Arrive}) and one action transition (\textit{Start}), which awards a reward every time an assignment is made. The top left triple presents, in order, the network tag (which has value $A$, meaning that only action transitions are enabled), the current time, and the cumulative reward. 

It may seem tempting to translate a marked A-E PN directly to a heterogeneous graph with places and transitions as (heterogeneous) nodes and edges corresponding to the A-E PN arcs. However, this approach proves problematic whenever a place contains more than one token since the corresponding node in the assignment graph should hold an embedding of the tokens, which is infeasible as their number is variable. Moreover, there would be no obvious way to encode the possible assignments. For the reasons above, we operate an intermediate step, namely, the expansion, to bring a given marked A-E PN to an equivalent representation suitable for translation to an assignment graph.

\begin{definition}[Expansion]
An expansion is a function from a source marked A-E PN $\mathcal{S}$ to a target marked A-E PN $\mathcal{T}$ that is functionally equivalent to $\mathcal{S}$ but contains at most one token in every place.
\end{definition}
In this work, we do not discuss the equivalence between source A-E PN and target A-E PN in terms of PN properties such as reachability or safety, since such properties are not relevant for the translation of the target A-E PN to the assignment graph.

At a high level, the expansion function operates in three main steps:
\begin{enumerate}
    \item \textbf{Place expansion}: each place in the source network, for each token in its marking, is expanded to a new place in the target network marked with that token. Empty places are expanded to a single empty place.
    \item \textbf{Transition expansion}: every evolution transition in the source network is expanded to a new evolution transition in the target network. Every action transition is expanded to a new action transition in the target network for each timed binding (a set of tokens enabling the original transition). 
    \item \textbf{Arc expansion}: new arcs are created in the target network connecting expanded places to expanded transitions, and vice-versa, obtained from places and transitions connected by an arc in the source network. 
\end{enumerate}

The expansion function is described in~\cref{alg:expansion}. In the algorithm, we use the subscript notation to refer to the individual elements of a given A-E PN. For example, if $S$ is a marked A-E PN, we refer to the set of places in $S$ as $P_S$. We will also use the notation $M_p$ to refer to the timed marking of place $p$. we use the notation $'$ to refer to the creation of new PN elements. We denote with $p(t)$ a place $p$ marked with a single token $t$. $M_p$ represents the marking of a place $p$. $b_{tr}$ is the timed binding of a transition $tr$. Moreover, we introduce an auxiliary multi-set $H$ to map places of $\mathcal{S}$ into places of $\mathcal{T}$.

\begin{algorithm}
\caption{A-E PN Expansion}
\label{alg:expansion}
\begin{algorithmic}[1]
\Procedure{expand}{$\mathcal{S}$}
    \State $A_{\mathcal{T}} \gets A_{\mathcal{S}}$, $P_{\mathcal{T}} \gets \emptyset$, $T_{\mathcal{T}} \gets \emptyset$, $\mathcal{A}_{\mathcal{T}} \gets \emptyset$, $H \gets \emptyset$
    \For{$p \in P_\mathcal{S}$}
        \If{$M_p \neq \emptyset$}
            \For{$t \in M_p$}
                
                \State $P_{\mathcal{T}} \gets P_{\mathcal{T}} \cup \{p'(t)\}$, $H \gets H \cup (p, p'(t))$
            \EndFor
        \Else
            \State $P_{\mathcal{T}} \gets P_{\mathcal{T}} \cup \{p\}$
        \EndIf
    \EndFor
    \For{$ tr \in T_{\mathcal{S}}$}
        \If{$L(tr) = \text{'E'}$}
            \State $T_{\mathcal{T}} \gets T_{\mathcal{T}} \cup \{tr\}$
            \For{$a \in F_{\mathcal{S}}$}
                \If{$a[0] = tr$}
                    \State $F_{\mathcal{T}} \gets F_{\mathcal{T}} \cup \{a\}$
                \EndIf
            \EndFor
        
        \Else
            \For{$b \in b_{tr}$}
                \State $tr' \gets tr$, $T_{\mathcal{T}} \gets T_{\mathcal{T}} \cup \{tr'\}$
                \For{$a \in F_{\mathcal{S}}$} 
                    \State $H' \gets {(p, p') \in H \text{ s.t. } p = a[1]}$, $P' \gets \{e[1] \forall e \in H' \}$ 
                    \If{$a[0] = tr$}
                        \State $F_{\mathcal{T}} \gets F_{\mathcal{T}} \cup {tr' \times P'}$
                    \Else \If{$a[1] = tr$}
                        \State $F_{\mathcal{T}} \gets F_{\mathcal{T}} \cup {P' \times tr'}$
                    \EndIf
                    \EndIf
                \EndFor
            \EndFor
        \EndIf
    \EndFor
    \State \textbf{return} $\mathcal{T}$
\EndProcedure
\end{algorithmic}
\end{algorithm}

\begin{figure}[h!]
    \centering
    \centerline{\includegraphics[width=0.96\textwidth]{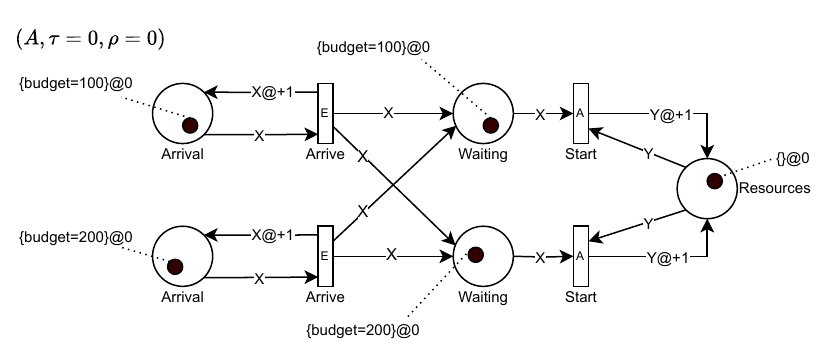}}
    \caption{A-E PN expanded from the A-E PN in~\cref{fig:example_problem}.}
    \label{fig:expansion_example}
\end{figure}

The output of the expansion function applied on the A-E PN in~\cref{fig:example_problem} is reported in~\cref{fig:expansion_example}.

The target A-E PN is easily translatable to a decision graph since every place now contains a single token, and every action transition in $\mathcal{T}$ corresponds to one enabling timed binding of the original action transition in $\mathcal{S}$. To operate the translation, we define the Petri Net to assignment graph mapping function.

\begin{definition}[Mapping from A-E PN to assignment graph]
A mapping from (expanded) A-E PN to assignment graph is a function from a marked A-E PN $\mathcal{T}$ to an assignment graph $G$ having one node for each place or transition in $\mathcal{T}$ and one (directed) edge for each arc in $\mathcal{T}$ to the corresponding source and destination nodes in $G$.
\end{definition}

At a high level, the mapping function operates in three main steps:
\begin{enumerate}
    \item \textbf{Place mapping}: every non-empty place is mapped to a node with the same attributes as the single token in the original place, plus its time.
    \item \textbf{Transition mapping}: every transition is mapped to a node without attributes.
    \item \textbf{Arc mapping}: every arc not connected to empty places is mapped to an edge connecting nodes corresponding to the arc's source and destination.
\end{enumerate}

The mapping algorithm is described in~\cref{alg:translation}. In the algorithm, we use the notation $v(t)$ to refer to the value of attributes of token $t$, $\tau (t)$ to refer to its time, and $v(n')$ to refer to the values associated with attributes of the type of the newly created node $n'$. Finally, we introduce two support multi-sets, $H_p$ and $H_t$, used to store tuples of nodes and transitions in the A-E PN and the corresponding edges in the assignment graph.

\begin{algorithm} \caption{Petri Net to Assignment Graph Mapping} \label{alg:translation} \begin{algorithmic}[1] \Procedure{map}{$\mathcal{T}$} 
\State $V_G \gets \emptyset$, $D_G \gets \emptyset$, $Y_G \gets \emptyset$, $D_G \gets \emptyset$, $A_G \gets A_T$, $H_p \gets \emptyset$, $H_t \gets \emptyset$
\For{$\forall p \in P_{\mathcal{T}}$}
    \If{$\exists p \text{ s.t. } p(t)$}
        \State $V_G \gets V_G \cup \{n' \text{ s.t. } \phi(n') = \{\text{TIME}\} \cup \mathcal{A}(p) \land v(n') = \{\tau (t)\} \cup v(t) \}$
        \State $H_p \gets H_p \cup (p, n')$
    \EndIf
\EndFor
\For{$\forall tr \in T_{\mathcal{T}}$}
    \If{$L_{\mathcal{T}}(tr) = \text{'E'}$}
        \State $V_G \gets V_G \cup {n' \text{ s.t. } \phi(n') = \text{E\_Transition} \land v(n') = \emptyset}$
    \Else
        \State $V_G \gets V_G \cup {n' \text{ s.t. } \phi(n') = \text{A\_Transition} \land v(n') = \emptyset}$
    \EndIf
    \State $H_{tr} \gets H_{tr} \cup (tr, n')$
\EndFor
\For{$\forall a \in F_T$}
    \If{$a[0] \in P_{\mathcal{T}}$}
        \State $D_G \gets D_G \cup \{(H_p[a[0]], H_{tr}[a[1]])\}$
    \Else
        \State $D_G \gets D_G \cup \{(H_{tr}[a[0]], H_p[a[1]])\}$
    \EndIf
\EndFor
\State \textbf{return} $G$
\EndProcedure
\end{algorithmic}
\end{algorithm}

\begin{figure}[h!]
    \centering
    \centerline{\includegraphics[width=0.9\textwidth]{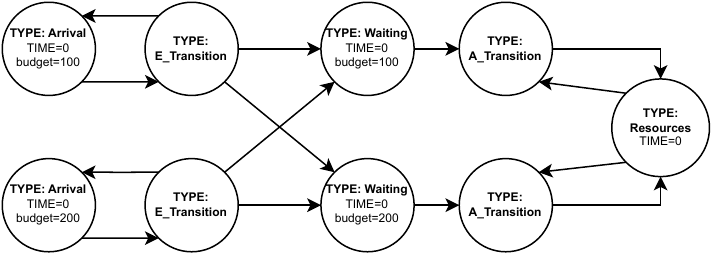}}
    \caption{Decision graph mapped from the A-E PN in~\cref{fig:expansion_example}.}
    \label{fig:mapping_example}
\end{figure}

The mapping function's output applied on the A-E PN in~\cref{fig:expansion_example} is reported in~\cref{fig:mapping_example}.

In an assignment graph, selecting a node corresponding to an action transition directly translates to choosing a binding to fire the transition. By following the expansion-mapping procedure, we obtain an assignment graph where each binding is represented as a node, effectively casting the task assignment problem to a node selection problem on the assignment graph. Moreover, the feature representation technique based on assignment graphs produces graph observations that encode the entirety of the information contained in a given PN marking. 
\section{Learning policies on graphs}\label{ppo_graph}
This section provides an overview of the steps necessary to train PPO using assignment graphs as observations. To the authors' knowledge, this work presents the first application that leverages fully dynamic graphs in PPO.

\begin{figure}[h!]
    \centering
    \centerline{\includegraphics[width=0.9\textwidth]{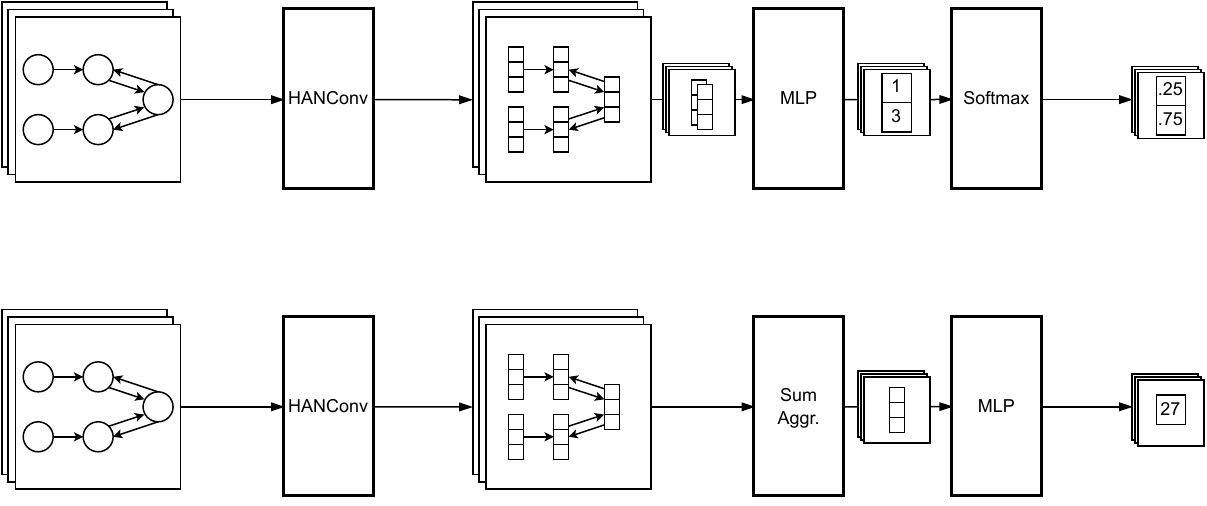}}
    \caption{Policy and Value Networks}
    \label{fig:networks}
\end{figure}

Decision graphs represent a flexible feature representation approach, but their shapes vary significantly from sample to sample. For this reason, we need NN layers capable of handling heterogeneous graphs of varying sizes seamlessly, such as the Heterogeneous Graph Attention Layer (HANConv)~\cite{wang2021heterogeneous}. In HANConv layers, information is shared across nodes of the same type and then across nodes of different types, effectively bringing each node to a fixed-size representation known as embedding. As shown in~\cref{fig:networks}, both the policy and the value networks employ the HANConv layer as the encoder. The two networks, however, differ in the decoding phase:
\begin{itemize}
    \item \textbf{Policy network decoder}: encoded action nodes are passed through a two-layer MLP with RELU activation function one by one, producing a value per each node as output. The outputs of the MLP are stacked in a vector that is then passed through a softmax layer that ensures they collectively sum up to $1$, effectively turning the MLP outputs into the probability each node has to be selected.
    \item \textbf{Value network decoder}: all encoded nodes are passed through an aggregation layer that computes their mean, and the resulting vector is passed to a two-layer MLP with RELU activation function, producing a single value as output.
\end{itemize}

However, choosing a suitable network architecture is not enough to handle assignment graphs in PPO. In fact, the dynamic nature of assignment graphs introduces two main challenges when performing batch learning:

\begin{enumerate}
    \item \textbf{Batches contain samples of different sizes}: every assignment graph can be of a different size, making the batching procedure complex. Instead of setting a maximum graph size $m$ and padding every observation to size $m$, we adopt the approach used in the \textit{Pytorch Geometric} library~\cite{DBLP:journals/corr/abs-1903-02428}, based on storing a large (sparse) adjacency matrix by concatenating those of single observations along the main diagonal. This approach is much more memory-efficient and allows for modeling assignment graphs of arbitrary sizes. 
    \item \textbf{The number of actions is variable}: in classic RL applications, all possible actions are listed at initialization. In the proposed method, actions correspond to $v_a$ nodes in the assignment graphs, and their number is variable from sample to sample. The use of HanConv ensures that the order of actions is irrelevant (permutation invariance), and the variable amount of actions can be handled using indexed softmax to ensure the action probabilities computed at the output of the actor are correctly associated with the observation that generated them.
\end{enumerate}

Having provided an overview of the PPO implementation used to handle assignment graphs, we can proceed with evaluating the proposed method.
\section{Evaluation}\label{use_cases}
This section is dedicated to the definition of a set of three archetypal problem instances of increasing complexity: one with finite and small state/action space, easily representable in vectorial form; one with finite but large state/action space, which proves hard to solve by resorting on vectorial observations; one with infinite state/action space, not representable in vectorial form. We show that PPO  can learn close-to-optimal assignment policies in all three cases.

\begin{figure}[h!]
    \centering
    \centerline{\includegraphics[width=0.96\textwidth]{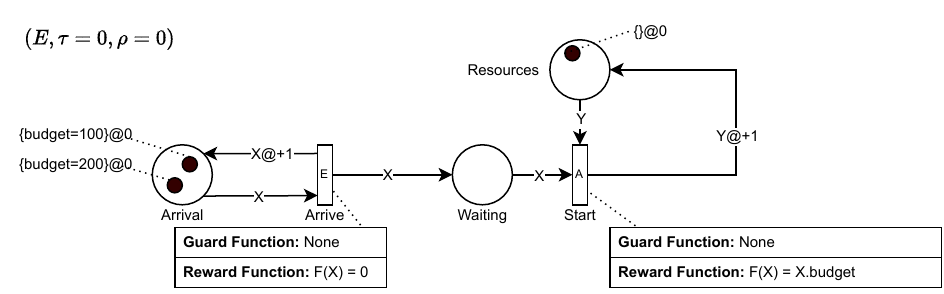}}
    \caption{Initial marking of the A-E PN representation of the first problem.}
    \label{fig:simple_problem}
\end{figure}

In~\cref{fig:simple_problem}, we report the initial configuration of a problem entailing a consulting company with a single employee deciding on which projects to take on, given that each project has a budget and the employee's objective is the maximization of the collected budget. In this case, the projects can only have a budget of $100$ or $200$, based on their type.

\begin{figure}[h!]
    \centering
    \centerline{\includegraphics[width=0.96\textwidth]{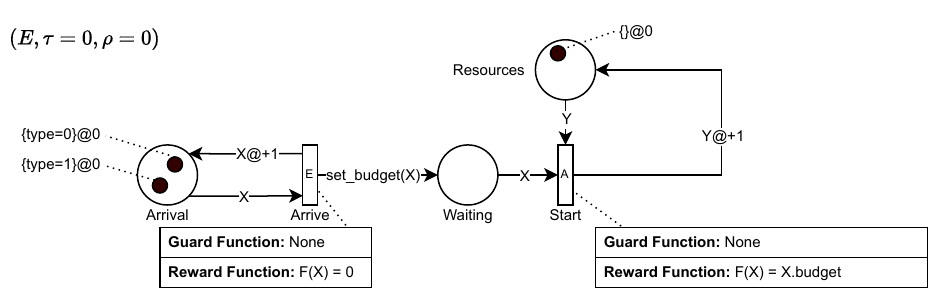}}
    \caption{Initial marking of the A-E PN representation of the second problem.}
    \label{fig:problem}
\end{figure}

In~\cref{fig:problem}, we report the initial configuration of a more complex problem where project budgets are values in euros sampled from uniform distributions between $70$ and $130$ for projects of type $0$ and between $170$ and $230$ for projects of type $1$, rounded to the second decimal value. In this case, the state and action spaces are still finite, but their dimensionality is very large.

\begin{figure}[h!]
    \centering
    \centerline{\includegraphics[width=0.96\textwidth]{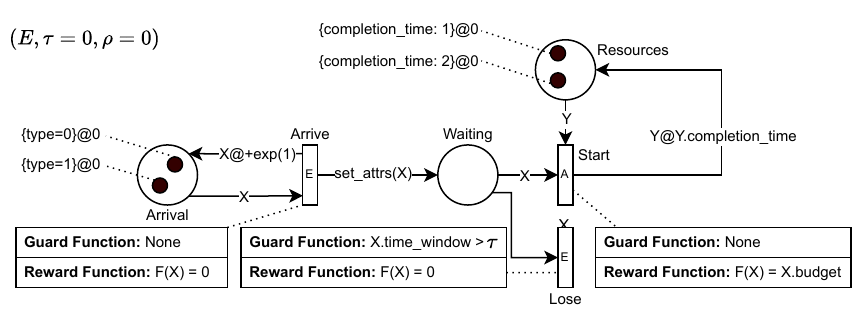}}
    \caption{Initial marking of the A-E PN representation of the third problem.}
    \label{fig:hard_problem}
\end{figure}

In~\cref{fig:hard_problem}, we report the initial configuration of a problem where the task budgets are sampled from a Gaussian distribution with mean respectively $100$ and $200$, depending on the task type, and standard deviation $10$. Tasks enter the system according to exponential interarrival times with rate $1$. New projects can only be accepted during a finite time window, which is a feature of the project and is sampled from an exponential distribution with rate $\frac{1}{2}$. Additionally, we consider two resources having different processing times. The Gaussian budgets and the time window features, set for each task through the \textit{set\_attrs} arc function, make the vectorial state and action spaces infinite.

The three problems were modeled and, where possible, solved with a PPO instance that uses vector observations (\textit{PPO Vector}) and one that uses assignment graphs (\textit{PPO Graph}). As a baseline, we provide the random policy (\textit{Random}) and a simple but effective greedy heuristic (\textit{Greedy}) that always prioritizes the tasks with the highest budget. Experiments were conducted on $10^5$ independent episodes, truncated after $10$ time units. The results are reported in~\cref{tab:results} with the mean and variance of the total reward accumulated in each episode. We include an indication of the order of magnitude of state and action space in the vectorial form.

\setlength{\tabcolsep}{5pt}
\begin{table}[]
\caption{Results for the presented problem instances.}
\label{tab:results}
\centering
\begin{tabular}{ll|llll} 
\toprule
\multicolumn{2}{c|}{Problem} & \multicolumn{4}{c}{Results} \\ 
\cmidrule{1-2} \cmidrule{3-6} 
Instance        & Size & Random                & Greedy & PPO Vector                  & PPO Graph \\ \midrule 
\cref{fig:simple_problem} & $10^0$ & \(1349 \pm 95\)  & \(2000 \pm 0\) & \(2000 \pm 0\)  & \(2000 \pm 0\) \\ 
\cref{fig:problem}     & $10^4$ & \(1348 \pm 109\) & \(1998 \pm 53\) & \(1625 \pm 130\) & \(1999 \pm 54\) \\ 
\cref{fig:hard_problem}   & $\infty$ & \(2096 \pm 231\)   & \(2266 \pm 316\) & Not representable  & \(2341 \pm 329\)  \\ 
\bottomrule
\end{tabular}
\end{table}

The experimental results outline how the assignment graph provides a suitable approach to feature representation regardless of the sizes of the action and state spaces. In the first problem instance, all solving methods reach perfect convergence. However, in the second instance, the PPO instance trained with vector observations struggles to converge, whereas the instance trained on assignment graphs performs the same as the heuristic. Even in the last problem, where the effect of random events is substantial, the PPO instance trained through assignment graphs matches the performance of the heuristic. All experiments are available in a proof-of-concept Python package\footnote{The code is publicly available at \url{https://github.com/lobiaminor/BPM2024}}.
\section{Conclusions and Future Work}\label{conclusion}

This paper proposed a universal feature representation approach for assignment problems, providing three contributions: the introduction of the assignment graph as a graph-based feature representation for assignment problems, the establishment of a mapping from marked CPN to assignment graphs, and the adaptation of the PPO algorithm to handle assignment graphs as observations.

These contributions were evaluated through their application to three archetypal assignment problems of increasing complexity. The results demonstrated that our method can learn close-to-optimal task assignment policies, regardless of the dimensionalities of the state and action spaces, where existing approaches were limited to state and action spaces of finite size.

Looking forward, the inherent scalability of the proposed universal feature representation approach shows promise. However, this work only focused on small-scale use cases, while its possible limitations are still to be fully explored. For this reason, future work will focus on applying the proposed methods to model and solve real-world problems.

\bibliographystyle{unsrt}
\bibliography{lnbib, newbib} 

\begin{thebibliography}{10}

\bibitem{10.1007/978-3-031-41620-0_13}
R.~Lo~Bianco, R.~Dijkman, W.~Nuijten, and W.~van Jaarsveld.
\newblock Action-evolution petri nets: A framework for modeling and solving dynamic task assignment problems.
\newblock In {\em Business Process Management}, pages 216--231, 2023.

\bibitem{zbikowski_deep_2023}
K.~{\.{Z}}bikowski, M.~Ostapowicz, and P.~Gawrysiak.
\newblock Deep reinforcement learning for resource allocation in business processes.
\newblock In {\em Process Mining Workshops}, pages 177--189, 2023.

\bibitem{middelhuis2024learning}
J.~Middelhuis, R.~Lo Bianco, E.~Scherzer, Z.~Bukhsh, I.~Adan, and R.~Dijkman.
\newblock Learning policies for resource allocation in business processes, 2024.

\bibitem{tavares2018algorithms}
Anderson~Rocha Tavares, Sivasubramanian Anbalagan, Leandro~Soriano Marcolino, and Luiz Chaimowicz.
\newblock Algorithms or actions? a study in large-scale reinforcement learning.
\newblock In {\em IJCAI}, pages 2717--2723, 2018.

\bibitem{routing_1980}
H.~N. Psaraftis.
\newblock A dynamic programming solution to the single vehicle many-to-many immediate request dial-a-ride problem.
\newblock {\em Transportation Science}, 14(2):130--154, 1980.

\bibitem{aggregation_survey}
D.~F. Rogers, R.~D. Plante, R.~T. Wong, and J.~R. Evans.
\newblock Aggregation and disaggregation techniques and methodology in optimization.
\newblock {\em Operations Research}, 39(4):553--582, 1991.

\bibitem{shape}
R.~Cheung and W.~B. Powell.
\newblock Shape - a stochastic hybrid approximation procedure for two-stage stochastic programs.
\newblock {\em Operations Research}, 48(1):73--79, 2000.

\bibitem{dynamic_assignment}
M.~Z. Spivey and W.~B. Powell.
\newblock The dynamic assignment problem.
\newblock {\em Transportation Science}, 38(4):399--419, 2004.

\bibitem{8786063}
G.~Park and M.~Song.
\newblock Prediction-based resource allocation using lstm and minimum cost and maximum flow algorithm.
\newblock In {\em 2019 International Conference on Process Mining (ICPM)}, pages 121--128, 2019.

\bibitem{10103533}
G.~Park and M.~Song.
\newblock Optimizing resource allocation based on predictive process monitoring.
\newblock {\em IEEE Access}, 11:38309--38323, 2023.

\bibitem{doi:10.1080/07421222.2002.11045693}
W.~M.P. Van Der~Aalst A.~Kumar and E.~M.W. Verbeek.
\newblock Dynamic work distribution in workflow management systems: How to balance quality and performance.
\newblock {\em Journal of Management Information Systems}, 18(3):157--193, 2002.

\bibitem{DASHTBOZORGI2023102198}
Z.~{Dasht Bozorgi}, I.~Teinemaa, M.~Dumas, M.~{La Rosa}, and A.~Polyvyanyy.
\newblock Prescriptive process monitoring based on causal effect estimation.
\newblock {\em Information Systems}, 116:102198, 2023.

\bibitem{9683135}
D.~Gammelli, K.~Yang, J.~Harrison, F.~Rodrigues, F.~C. Pereira, and M.~Pavone.
\newblock Graph neural network reinforcement learning for autonomous mobility-on-demand systems.
\newblock In {\em 2021 60th IEEE Conference on Decision and Control (CDC)}, pages 2996--3003, 2021.

\bibitem{YAN2023103260}
Y.~Yan, Y.~Deng, S.~Cui, Y.~Kuo, A.~Chow, and C.~Ying.
\newblock A policy gradient approach to solving dynamic assignment problem for on-site service delivery.
\newblock {\em Transportation Research Part E: Logistics and Transportation Review}, 178:103260, 2023.

\bibitem{1542cc6b364e4cf8b7c2adb51da444ce}
L.~Begnardi, H.~Baier, {W.} {van Jaarsveld}, and Y.~Zhang.
\newblock Deep reinforcement learning for two-sided online bipartite matching in collaborative order picking.
\newblock In {\em Proceedings of the 15th Asian Conference on Machine Learning (ACML2023)}, Proceedings of Machine Learning Research, 2023.

\bibitem{10.1007/978-3-030-98581-3_9}
A.~Chiorrini, C.~Diamantini, A.~Mircoli, and D.~Potena.
\newblock Exploiting instance graphs and graph neural networks for next activity prediction.
\newblock In {\em Process Mining Workshops}, pages 115--126, Cham, 2022. Springer International Publishing.

\bibitem{10.1007/978-3-030-94343-1_3}
S.~Weinzierl.
\newblock Exploring gated graph sequence neural networks for predicting next process activities.
\newblock In {\em Business Process Management Workshops}, pages 30--42, Cham, 2022. Springer International Publishing.

\bibitem{sommers2021process}
D.~Sommers, V.~Menkovski, and D.~Fahland.
\newblock Process discovery using graph neural networks, 2021.

\bibitem{HU20201}
L.~Hu, Z.~Liu, W.~Hu, Y.~Wang, J.~Tan, and F.~Wu.
\newblock Petri-net-based dynamic scheduling of flexible manufacturing system via deep reinforcement learning with graph convolutional network.
\newblock {\em Journal of Manufacturing Systems}, 55:1--14, 2020.

\bibitem{schulman2017proximal}
J.~Schulman, F.~Wolski, P.~Dhariwal, A.~Radford, and O.~Klimov.
\newblock Proximal policy optimization algorithms, 2017.

\bibitem{schulman2018highdimensional}
J.~Schulman, P.~Moritz, S.~Levine, M.~Jordan, and P.~Abbeel.
\newblock High-dimensional continuous control using generalized advantage estimation, 2018.

\bibitem{wang2021heterogeneous}
X.~Wang, H.~Ji, C.~Shi, Bai Wang, Peng Cui, P.~Yu, and Yanfang Ye.
\newblock Heterogeneous graph attention network, 2021.

\bibitem{DBLP:journals/corr/abs-1903-02428}
M.~Fey and J.~E. Lenssen.
\newblock Fast graph representation learning with pytorch geometric.
\newblock {\em CoRR}, abs/1903.02428, 2019.

\end{thebibliography}

\end{document}